# Integrating Non-Linear Radon Transformation for Diabetic Retinopathy Grading

Farida Mohsen[1,*], Samir Belhaouari[1,*], and Zubair Shah[1,*]

[1]College of Science and Engineering, Hamad Bin Khalifa University, Doha, Qatar

*Correspondence to: zshah@hbku.edu.qa

**ABSTRACT**

Diabetic retinopathy is a serious ocular complication that poses a significant threat to patients' vision and overall health. Early detection and accurate grading are essential to prevent vision loss. Current automatic grading methods rely heavily on deep learning applied to retinal fundus images, but the complex, irregular patterns of lesions in these images, which vary in shape and distribution, make it difficult to capture the subtle changes. This study introduces RadFuse, a multi-representation deep learning framework that integrates non-linear RadEx-transformed sinogram images with traditional fundus images to enhance diabetic retinopathy detection and grading. Our RadEx transformation, an optimized non-linear extension of the Radon transform, generates sinogram representations to capture complex retinal lesion patterns. By leveraging both spatial and transformed domain information, RadFuse enriches the feature set available to deep learning models, improving the differentiation of severity levels. We conducted extensive experiments on two benchmark datasets, APTOS-2019 and DDR, using three convolutional neural networks (CNNs): ResNeXt-50, MobileNetV2, and VGG19. RadFuse showed significant improvements over fundus-image-only models across all three CNN architectures and outperformed state-of-the-art methods on both datasets . For severity grading across five stages, RadFuse achieved a quadratic weighted kappa of 93.24%, an accuracy of 87.07%, and an F1-score of 87.17%. In binary classification between healthy and diabetic retinopathy cases, the method reached an accuracy of 99.09%, precision of 98.58%, and recall of 99.64%, surpassing previously established models. These results demonstrate RadFuse's capacity to capture complex non-linear features, advancing diabetic retinopathy classification and promoting the integration of advanced mathematical transforms in medical image analysis. The source code will be available at https://github.com/Farida-Ali/RadEx-Transform/tree/main

## Introduction

Diabetic retinopathy (DR) is a microvascular complication of diabetes mellitus and a leading cause of vision impairment and blindness among working-age adults worldwide. According to the International Diabetes Federation, the global prevalence of diabetes is projected to rise from 463 million in 2019 to 700 million by 2045, intensifying the burden of DR on healthcare systems[1]. DR is characterized by progressive damage to the retinal microvasculature, starting with microaneurysms and hemorrhages and advancing to proliferative diabetic retinopathy (PDR), where abnormal neovascularization can cause severe vision loss[2].

Early detection of DR is critical, as treatments are more effective before significant damage occurs. However, DR often progresses asymptomatically, and by the time symptoms are noticeable, considerable retinal damage has already taken place[3]. Hyperglycemia in diabetic patients damages retinal blood vessels, leading to leakage and the formation of exudates, hemorrhages, and microaneurysms. The extent and severity of these lesions are used to grade DR[2]. Despite the need for early detection, manual DR grading is labor-intensive, subject to inter-clinician variability, and requires specialized expertise[4].

Automated DR grading using deep learning offers a promising solution to these limitations. Convolutional neural networks (CNNs) have demonstrated significant potential in medical image analysis, particularly for DR detection and classification[5,6]. However, relying solely on retinal fundus images presents inherent limitations. CNN-based models often face challenges in capturing fine-grained retinal features, especially in early-stage DR, where lesions like microaneurysms and small hemorrhages are subtle and less pronounced[7]. These lesions are difficult to detect due to their small size, irregular shape, and sparse distribution within the retina[7–9]. The complex, non-linear nature of retinal lesions, combined with the spatial variability of their distribution, makes it challenging for CNNs to extract detailed features and establish connections between different lesion regions[10–13]. Furthermore, adjacent DR stages often exhibit minimal visual differences, leading to difficulties in accurately distinguishing between them[11,13].

To address these challenges, researchers have explored the integration of advanced image transformations with deep learning models. The Radon transform, which computes projections of an image along various angles, has been used successfully in medical imaging, particularly in computed tomography, for image reconstruction and feature extraction[14,15]. Its key benefit is the ability to simplify a complex image structure into analyzable projections. By projecting image data along linear paths, the Radon transform can highlight critical structures and edges. For example, incorporating Radon transforms has improved feature extraction in tumor detection by highlighting linear structures and edges[15]. Tavakoli et al.[12] demonstrated the Radon transform's effectiveness in enhancing microaneurysm detection in retinal images. Another study by Raaj et al.[16] applied the Radon transform to mammogram images to classify them into normal, benign, and malignant categories, achieving high performance using hybrid CNN architecture. However, the linear Radon transform struggles to capture non-linear features such as curved edges or irregular textures, typical in pathological conditions. Its linear assumptions may not accurately represent these complex features, potentially leading to missed diagnoses or inaccuracies. To overcome these limitations, we previously introduced the RadEx transformation, a non-linear extension of the Radon transform, designed to capture non-linear and complex subtle features in image data[17]. While the RadEx transformation has been applied to chest X-rays for COVID-19 detection, its use in retinal image analysis for DR grading remains unexplored.

In this study, we propose RadFuse, a novel multi-representation deep learning approach that integrates RadEx-based sinogram images with original fundus images, creating a robust multi-representation input and providing complementary perspectives for DR detection and severity grading. Although both image types originate from the same fundus image, the RadEx transformation generates a distinct feature space that complements the spatial information from the original images to capture non-linear lesion patterns and distributions that are not readily discernible in the raw images. The rationale behind generating and including RadEx-transformed images as additional input is to capture intricate, non-linear lesion patterns that are often subtle and distributed across the retinal surface. While CNNs are proficient at extracting diverse feature types, the complex, non-linear nature and spatial variability of retinal lesions present challenges for standard architectures when processing retinal disease images alone[10]. The RadEx transformation serves as a secondary representation, enriching the feature set and providing an additional layer of diagnostic information, thereby enhancing the model's ability to detect subtle and complex lesion patterns that may be overlooked by single-representation models. This dual-representation approach enables the model to leverage complementary features, improving its ability to detect subtle and distributed lesions essential for accurate DR grading.

This approach is supported by previous studies[12,16], which demonstrated that transformations like the Radon transform emphasize unique features not easily discernible in the original spatial representations. These transformations provide a complementary view that, when combined with the original images, enables a more comprehensive analysis. The RadEx transformation, as a non-linear extension of the Radon transform, further enhances the model's ability to capture complex, distributed lesion patterns in retinal images, which are critical for accurate DR grading. Incorporating multi-modal inputs or transformed images, therefore, provides complementary information, capturing lesion diversity more effectively and improving diagnostic accuracy[10].

To the best of our knowledge, this is the first study to use a multi-representation approach combining non-linear RadExtransformed sinogram images with retinal fundus images for DR grading. These sinogram representations highlight non-linear features associated with DR and provide additional information that enriches the deep learning model, significantly boosting CNN performance in detection and grading. The main contributions of this work are:

- Innovative multi-representation approach utilizing non-linear RadEx transformation: We optimized and applied the nonlinear RadEx transformation to retinal images, generating sinograms as a new representation of the data . These sinograms serve as an additional representation, offering a complementary perspective to traditional retinal images for DR detection and severity grading . This transformation detects non-linear features and microvascular abnormalities associated with different DR severity levels. We then fused the sinogram and fundus images to constitute a multi-representation input, enabling the model to learn complementary features from both representations (see Section ).

- Comprehensive evaluation and validation of the proposed approach on two benchmark datasets, APTOS-2019 and DDR : We evaluated our approach on the Asia Pacific Tele-Ophthalmology Society (APTOS) 2019 dataset, which was released as part of the Kaggle blindness detection challenge[18], and the DDR dataset[19] which is the second-largest publicly available dataset for DR grading. Extensive experiments, including comparisons using retinal image-only, RadEx-only, and multi-representation images with three different CNN architectures, were conducted to validate the effectiveness of our proposed approach. Our method demonstrated significant improvements over retinal image-only models and outperformed existing state-of-the-art (SOTA) methods on both datasets.

Our findings indicate that integrating non-linear Radon transformations provides an effective means of capturing complex non-linear features in retinal images, leading to accurate DR severity grading. This work advances the current state of DR classification and opens up new possibilities for integrating advanced mathematical transforms into medical image analysis pipelines.

## Methods

This section outlines our proposed multi-representation deep-learning approach for DR detection and severity grading. The approach consists of two primary components: the generation of RadEx-sinogram images and the implementation of multirepresentation deep learning.

**Proposed Approach**
*RadEx Transformation: Theoretical Overview*
The non-linear RadEx transformation is an advanced extension of the traditional Radon transform, designed specifically to capture complex, non-linear patterns within medical images[17]. Unlike the Radon transform, which relies on straight-line projections, the RadEx transformation employs parameterized curves that adapt to non-linear structures in the image. This allows for a more refined extraction of features critical to detecting and grading diabetic retinopathy, such as curvilinear blood vessels and scattered exudates. The RadEx transformation is defined by the following equation:

$$= \frac{M}{2} \left( z \frac{e^{c(p-q)} - 1}{e^{c(b-q)} + 1} + 1 \right) \quad (1)$$

where $z$ is the transformed vertical coordinate, $p$ is the horizontal coordinate, $M$ is the image size (assuming a square image of dimensions $M \times M$), The parameters $q$ and $c$ control the horizontal shift and the curvature of the transformation, respectively.

Although the original RadEx transformation shows potential, it faces limitations when applied to high-resolution retinal fundus images as demonstrated in Figure 1. This becomes especially problematic with larger images (e.g., 512x512 pixels and above), where significant regions of the image may remain underrepresented due to suboptimal parameter selection. In the context of DR grading, where lesions are distributed unevenly across the retina, the failure to cover these non-uniformly distributed areas poses a limitation to achieving high accuracy. Therefore, an optimized version of RadEx is necessary to ensure full coverage of the retinal image and enhance the detection of subtle, non-linear, and distributed lesions.

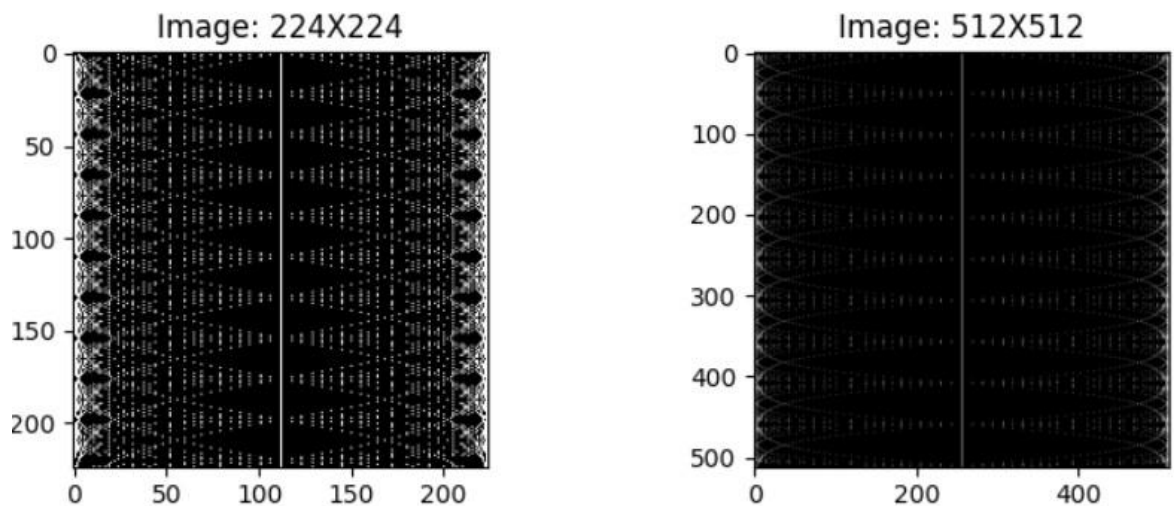

**Figure 1.** Visualization of pixel coverage by the RadEx transform across different image resolutions. Blank images of identical dimensions are used, with activated pixels representing areas covered by the transform. Black areas indicate uncovered regions.

*Optimization of the RadEx Transformation*
To ensure full image coverage and enhance feature extraction, we propose an optimized RadEx transformation focused on refining the selection of parameters $q$ and $c$. This optimization improves the transformation's ability to cover the entire image uniformly, reduce sparsity, and better capture critical retinal structures.

Selection of $q$ Values (Horizontal Shift): The parameter $q$ acts as a horizontal shift, determining the location of transformation curves across the image width. Extreme values of $q$ (either too negative or positive) can cause the curves to cluster at the image's edges, resulting in redundant information and inadequate coverage. To avoid this, we select $q$ values at regular intervals to ensure uniform sampling across the entire image width. The $q$ values are computed as:

$$q_i = i \times \Delta q \quad \text{for} \quad i = 0, 1, \ldots, m$$

where $\Delta q$ represents the step size, calculated as $\Delta q = \frac{M}{m}$, where $M$ is the image dimension (512 pixels in our case), and $m$ is the number of increments or divisions along the $q$ axis. As shown in Figure 2 (left), when $q$ values are spaced with increments of approximately 10 pixels ($\Delta q \approx 10$), the resulting curves demonstrate that the entire width of the image is covered uniformly. For comparison, Figure 2 (right) illustrates the curves when larger increments of approximately 44 pixels are used ($\Delta q \approx 44$), resulting in less frequent coverage across the image width. This comparison reinforces the need for appropriate selection of $m$ to ensure adequate coverage of the image. For this study, we set $m = 50$, resulting in $\Delta q = \frac{M}{50} \approx 10$. This choice was made after conducting a sensitivity analysis of different step sizes to evaluate the transformation behavior across various settings and select the optimal one that balances model performance and computational cost.

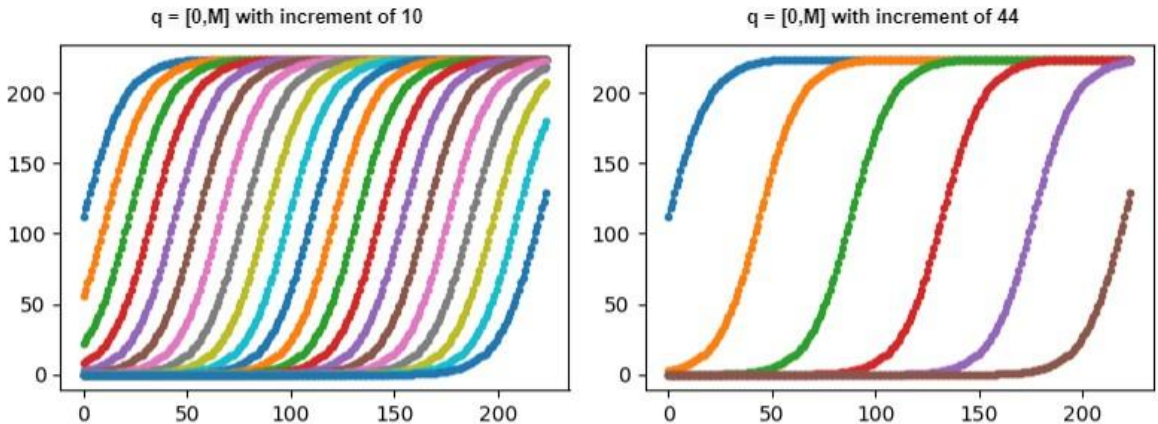

**Figure 2.** Plots showing pixel selections for varying $q$ values from 0 to $M$ at different increments, keeping $c$ constant. The X-axis represents the image width, and the Y-axis represents the image height.

Selection of $c$ Values (Curvature Control): The parameter $c$ controls the curvature of the transformation curves. Excessively large or small $c$ values cause the curves to quickly reach the image edges, potentially missing important regions. To avoid this, we constrain $c$ within a moderate range ($-1 \leq c \leq 1$) and use gradual increments to achieve a balanced distribution of curves across the image (see Figure 3). This ensures more uniform coverage and improves pixel selection granularity, which is crucial for detecting subtle lesions. We reformulate the RadEx transformation to precalculate $c$ for each $z$ using the inverse RadEx equation:

$$c = \frac{1}{p - qM} \log\left(\frac{z}{-z}\right)$$

This reformulation ensures that the transformation spans the full vertical range of the image, capturing essential features across all regions. We conclude that the optimal count of specialized $c$ values should approximate half the image size to achieve nearly 99% coverage, as shown in Figure 4, where the count of specialized $c$ values is $M/2$. Algorithm 1 outlines the steps for applying the transformation and generating RadEx-transformed images (sinograms), which are essential for our multi-representation deep learning model.

**Image Preprocessing and RadEx Sinogram Generation**

Several preprocessing steps were performed before applying the RadEx transformation to standardize the images and enhance feature visibility. First, we cropped the images to remove the black background, focusing solely on the retinal region. We then employed the Ben-Graham preprocessing method, a widely used technique in retinal image analysis[20]. This method involves multiple stages aimed at standardizing and improving image quality. The images were resized to a consistent resolution to ensure uniformity across the dataset. Following this, pixel intensity values were normalized to standardize brightness and contrast, reducing variability caused by different imaging conditions and equipment. A Gaussian blur was also applied to minimize noise while preserving important retinal structures.

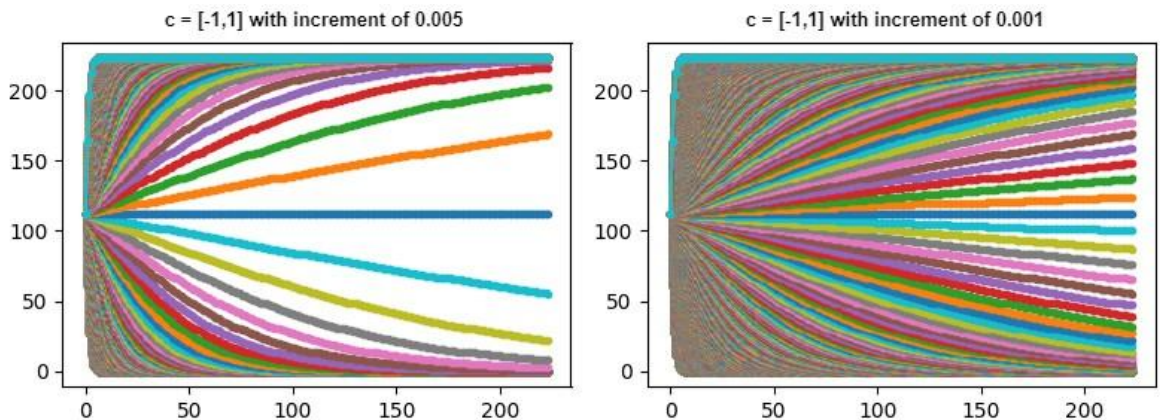

**Figure 3.** Plotting the pixels chosen for changing values of $c$ between -1 and 1 at different intervals, keeping $q$ constant at zero. The X-axis depicts image pixels in the horizontal direction while the y-axis depicts image pixels in the vertical direction.

After image preprocessing, we generated sinogram representations of the fundus images using our optimized RadEx transformation. For the optimized RadEx transformation, $q$ values are selected uniformly between 0 and $M$, with increments of $M/50$. For each $q$, we use $M/2$ values of $c$ (e.g., 112 for $M$ = 224) to ensure dense vertical coverage. The resulting sinograms were then normalized to ensure compatibility with the original images during fusion. Figure 5 illustrates a comparison between RadEx and traditional linear Radon-based sinograms across different DR grades. The RadEx-sinograms provide enhanced feature visibility, making subtle variations more distinguishable compared to linear Radon sinograms. Moreover, the differentiation between DR grades is more pronounced in the RadEx-sinograms, facilitating more accurate classification of severity levels by deep learning models.

The integration of these sinograms with the original fundus images forms the basis of the multi-representation (multimodal) input framework. This framework combines the non-linear, lesion-focused representation provided by the RadEx-transformed sinograms with the spatial details inherent in the original images. Together, these inputs enrich the feature space available to deep learning models, enabling better feature extraction. This dual-input approach is particularly valuable for early disease's detection, where subtle indicators carry critical diagnostic and prognostic significance.

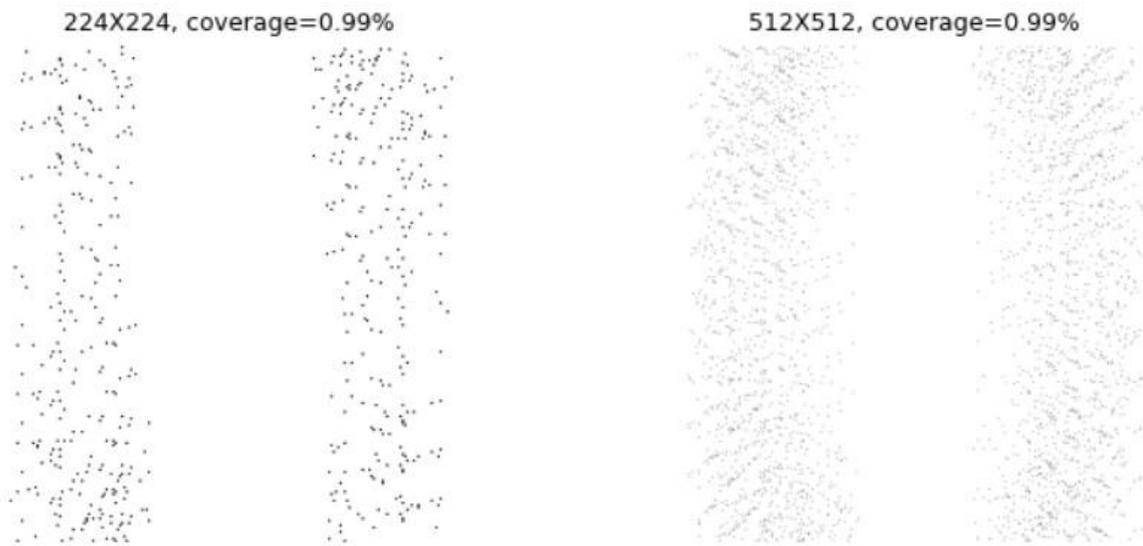

**Figure 4.** Visualization of pixel coverage by the optimized RadEx transform for different image sizes when the number of *c* is $M/2$. Blank images of identical dimensions are used, with activated pixels representing areas covered by the transform. White areas indicate covered regions.

---

Algorithm 1 Optimized RadEx Image Transformation

---

1: $M \leftarrow$ Size of image  2: *count_special_c* $\leftarrow \frac{M}{2}$  ▷ Number of special *c* values

3: Initialize *c_list* as an empty list

4: *q_values* ← sequence from 0 to $M$ with step $\frac{M}{m}$

5: $p \leftarrow \frac{M}{2}$

6: for *q* in *q_values* do

7:     for $z \leftarrow \frac{M}{2}$ to $M$ do

8:         *c* ← compute_c_given(*p*, *z*, *q*, *M*)

9:         Append *c* to *c_list*

10:    end for

11: end for

12: Sort *c_list*

13: *c_neg_list* ← range(−1, min(*c_list*), 0.1)

14: *c_pos_list* ← range(max(*c_list*), 1, 0.1)

15: *c_list_chosen* ← select *count_special_c* elements from *c_list*

16: *cs* ← concatenate([*c_neg_list*, *c_list_chosen*, *c_pos_list*]) 17:
for *q* in *q_values* do

18:    for *c* in *cs* do

19:       *p_array* ← [*i* for *i* in range(*M*)]

20:       *z_array* ← compute_z_given(*p_array*, *q*, *c*, *M*)

21:       *z_array* ← bound(*z_array*, max value = *M*)

22:    end for                     ▷ Ensures *z* values are between 0 and *M*

23: end for

**Multi-Representation Deep Learning Framework**

We used three CNN architectures to evaluate the efficacy of our multi-representation approach: ResNeXt-50[21], MobileNetV2[22], and VGG19[23]. ResNeXt-50 is known for its strong feature extraction capabilities through its grouped convolution strategy, while MobileNetV2 is optimized for efficient performance, making it suitable for deployment in resource-constrained environments. VGG19, although deeper and more computationally demanding, has demonstrated high accuracy in various image classification tasks.

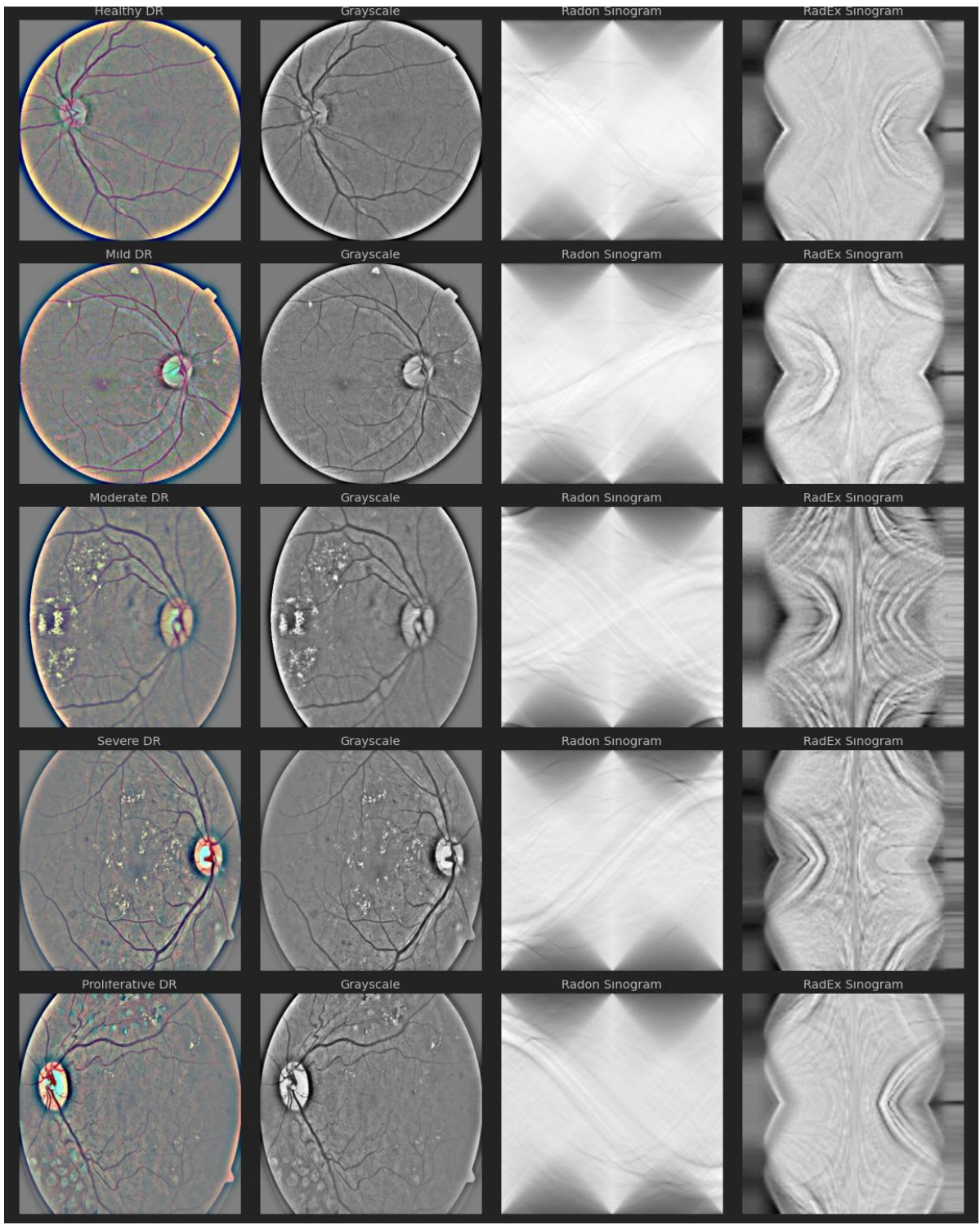

**Figure 5.** This figure illustrates the progression of different representations of retinal images for various DR stages: Healthy, Mild, Moderate, Severe, and PDR. The first column shows the original color fundus images. The second column presents grayscale versions of these images before applying RadEx. The third column contains linear Radon transform sinograms. The final column displays sinograms generated using our RadEx transformation..

    To maximize the potential of our proposed RadEx transformation, we integrated the sinogram representations with the original fundus images (RadFuse) through an early fusion strategy[24]. In this approach, the RadEx-transformed sinogram and the original image were horizontally concatenated to form a multi-representation input, allowing the CNNs to process both

the spatial and transformed-domain features simultaneously ( Fig. 6). This multi-representation framework enables the network to capture complementary information from both domains, thus enhancing feature representation and improving the classification of DR severity. Additionally, we conducted experiments on two benchmark datasets—APTOS-2019 and DDR—to validate the generalizability and robustness of our RadFuse framework. These datasets provide diverse retinal image samples, ensuring that our approach performs consistently across different datasets.

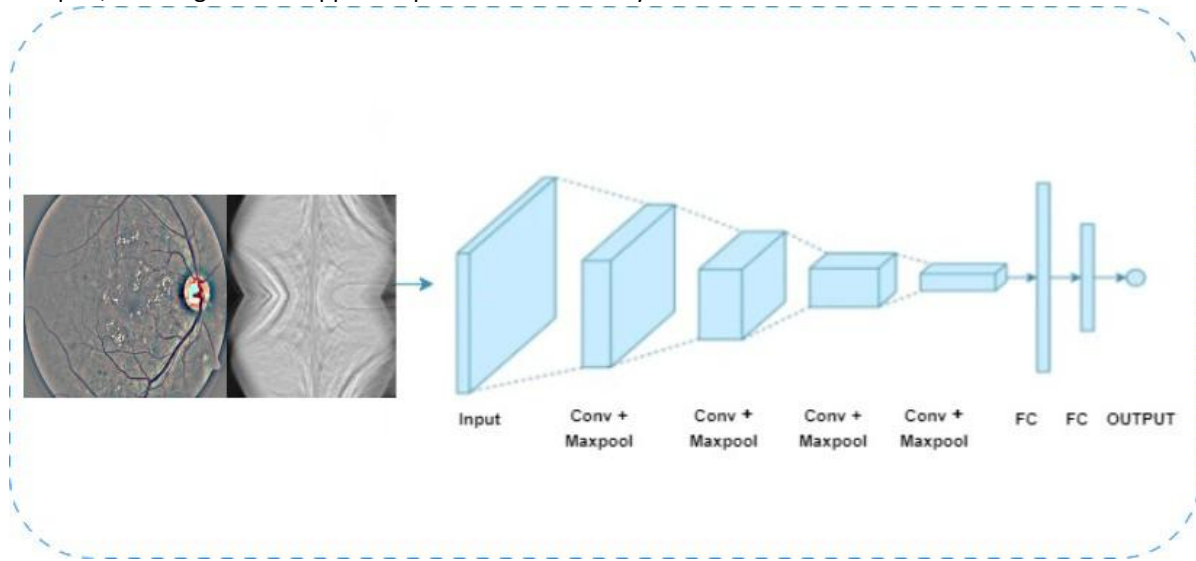

**Figure 6.** Our proposed multi-representation RadFuse framework for DR detection and grading. The input comprises two images: the original retinal fundus image and the RadEx-sinogram image. These two modalities are concatenated and processed through the CNN architecture. The final output layer utilizes a sigmoid activation function for binary DR detection and a SoftMax activation function for multi-class DR severity grading. This multi-representation approach enables the model to leverage both spatial and RadEx-transformed features, improving the accuracy and robustness of DR detection and severity classification.

***Time Complexity***

The time complexity of the optimized RadEx transformation algorithm can be analyzed based on the following factors:

- Outer Loop Over *q* Values: The loop runs from 0 to *M* with a step of $\frac{M}{10}$. Hence, it runs 10 iterations (since the step size is $\frac{M}{10}$ and the total range is *M*).

- Inner Loop Over *z* Values: For each value of *q*, the inner loop runs from $\frac{M}{2}$ to *M*, resulting in $\frac{M}{2}$ iterations.

- Computing *c* Values: In each iteration of the inner loop, a *c* value is computed, resulting in $\frac{M}{2}$ computations for each *q* value.

- Sorting and Range Operations for *c* Values: Sorting the *c* list takes $O(n \log n)$ time complexity, where *n* is the size of the list, approximately $10 \times \frac{M}{2} = 5M$.

- Nested Loops Over *q* and *cs* Values: For each *q* (10 iterations), the loop over *cs* runs. The length of *cs* is determined by the number of selected *c* values, which is approximately $\frac{M}{2}$.

Thus, the dominant factor in the time complexity is the nested loops over *q* values and *cs*, resulting in a time complexity of approximately $O(10 \times \frac{M}{2} \times M) = O(5M^2)$. Therefore, the overall time complexity is approximately $O(M^2)$, considering the significant computational steps involved.

## Experiments and Results

### Datasets

To evaluate the robustness and generalizability of the proposed RadFuse framework, we conducted experiments on two distinct datasets: APTOS-2019 and DDR.

APTOS-2019 Dataset[18]: This dataset, released as part of the 2019 Kaggle blindness detection challenge , contains 3,662 high-resolution color fundus images captured with various clinical cameras in controlled labe nvironments. The images in the dataset are categorized into five stages of DR: no DR (0), mild (1), moderate (2), severe (3), and PDR (4). The imagese were divided into training (80%), and testing (20%) sets for diagnosing and grading DR stages. As the data is highly unbalanced, we slected 10% of the training set for validation. To align with established literature, we designed the test set to include 20% of each category from the original APTOS-2019 dataset. This approach allowed for direct comparison with previous studies that used a similar evaluation fraction. The same test data was employed across all experiments to ensure consistency in evaluations. The distribution of images for both binary and multi-class classification tasks is detailed in Table 1.

DDR Dataset[19]: This dataset is the second-largest publicly available dataset, comprising 13,673 fundus images divided into 6,835 training, 2,733 validation, and 4,105 test images. Each image is graded into six categories by seven trained graders based on the International Classification of DR. Images of poor quality without clearly visible lesions are labeled as ungradable. Thus, the six levels are: no DR, mild DR, moderate DR, severe DR, PDR, and ungradable. For our experiments, we focused on the five-class DR grading task, excluding ungradable images. Consequently, our final dataset included 6,320 training, 2,503 validation, and 3,759 test images. The distribution of the dataset is imbalanced in that the normal images are more than the DR images. To ensure consistency, we applied the same preprocessing steps and model configurations as with the APTOS-2019 dataset.

| Classification | DR Stage | Number of images | Train | Validation | Test |
| --- | --- | --- | --- | --- | --- |
| Multi-class | No DR | 1805 | 1264 | 180 | 361 |
|  | Mild | 370 | 259 | 37 | 74 |
|  | Moderate | 999 | 701 | 99 | 199 |
|  | Severe | 193 | 136 | 19 | 38 |
|  | PDR | 295 | 207 | 29 | 59 |
| Binary | No DR | 1805 | 1264 | 180 | 361 |
|  | DR | 1857 | 1303 | 184 | 370 |

**Table 1.** Proportion of images in the training, validation, and test sets for multi-class and binary classification of DR stages.

### Experimental Setup

We implemented our proposed strategy using PyTorch 2.1.2 on a Linux operating system, leveraging an Nvidia GeForce RTX 2080 Ti GPU for both training and testing. The AdamW optimizer was used with a learning rate of $1\times10^{-4}$, which includes weight decay to improve generalization. A batch size of 16 was used for all experiments, and images were resized to 512×512 pixels for both the training and testing phases. The models, initialized with ImageNet pre-trained weights, were fine-tuned on the APTOS-2019 and DDR datasets. The training spanned 100 epochs, with early stopping applied after 10 epochs to prevent overfitting. Cross-entropy loss was the loss function, and data augmentation included blurring, flipping (vertical and horizontal), random rotation, sharpening, and adjustments to brightness and contrast. For our multi-representation model, we first concatenate the fundus image and the RadEx-transformed image to create a composite image. Then, augmentations are applied to the composite image.

### Evaluation Metrics

The imbalanced nature of the datasets presents challenges when using accuracy as a performance metric, as it tends to favor the majority class, often at the expense of the minority class. To overcome this, We prioritize the Quadratic Weighted Kappa (QWK) score as our primary evaluation metric to overcome this. QWK is designed to measure inter-rater agreement in multi-class classification by comparing expected and predicted scores, with values ranging from -1 (indicating complete disagreement) to
1 (indicating perfect agreement). A higher QWK score reflects stronger model performance and agreement with true labels. In addition to QWK, we report other key metrics such as Area Under the Curve (AUC), F1-score, recall, precision, accuracy, and Matthews correlation coefficient (MCC), ensuring a comprehensive evaluation of the model's performance, particularly in handling imbalanced data.

**Comparative Experiments**

This section evaluates the effectiveness of our RadFuse approach on both the APTOS-2019 and DDR datasets on the five-class severity grading, a clinically relevant and more challenging task. We used three CNN architectures—ResNeXt-50, MobileNetV2, and VGG19- to ensure the improvements are not architecture-dependent, as described in the Methods section. We first compare our RadFuse results with image-only, and RadEx-only models and then benchmark against several state-of-the-art (SOTA) models on both datasets. All experiments followed the same setup and data augmentation techniques for fair comparison.

*Results on APTOS-2019 dataset*

For severity grading, Table 2 shows the performance of our RadFuse models compared to the image-only and RadEx-only configurations across multiple metrics. RadFuse consistently showed improved performance across all three architectures: ResNeXt-50, MobileNet, and VGG-19. Notably, the ResNeXt-50-based RadFuse model achieved the highest metrics, with a QWK of 93.24%, outperforming the image-only (90.62%) and RadEx-only (78.78%) models. It also attained an AUC of 96.41%, and an F1-score of 87.17%, significantly outperforming its image-only and RadEx-only counterparts. Moreover, the RadFuse models showed a significant increase in MCC across all architectures compared to image-only and RadEx-only models. For example, the ResNeXt-50 RadFuse model achieved an MCC of 80.50%, an improvement of 5.96 percentage points over the image-only model. This substantial increase indicates that the RadFuse model is more effective in handling class imbalance and accurately detecting underrepresented classes. VGG19 with RadFuse also showed enhanced performance,
with the MCC increasing from 76.55 %, in the image-only model to 79.4%, thereby improving the DR classification accuracy. MobileNetV2 showed more modest gains but still improved with the RadFuse approach, indicating the consistent benefits of the multi-representation approach.

RadEx-only models performed moderately well but did not reach the high-performance levels of image-only or RadFuse models, indicating that while RadEx transformation captures valuable features, these are most effective when combined with original images. The improvements across various CNN architectures underscore the reliable advantages of integrating RadEx-transformed sinogram images with traditional fundus images within the RadFuse framework..

**Table 2.** Comparison of model performance for severity-level grading using different configurations. The best results are in bold. Image-only denotes models trained and tested using only fundus images, RadEx-only uses solely RadEx-transformed images, and RadFuse represents multi-representation models integrating both fundus images and RadEx-transformed images.

| Model | Configuration | QWK (%) | AUC | F1 (%) | MCC (%) | Accuracy (%) | Recall (%) | Precision (%) |
|---|---|---|---|---|---|---|---|---|
| VGG19 | RadFuse | 91.90 | 0.957 | 86.15 | 79.40 | 86.5 | 86.4 | 85.9 |
|  | Image-only | 90.80 | 0.950 | 84.19 | 76.55 | 84.4 | 84.6 | 83.8 |
|  | RadEx-only | 70.99 | 0.876 | 75.11 | 61.66 | 75.12 | 75.12 | 75.1 |
| MobileNetV2 | RadFuse | 90.91 | 0.954 | 86.11 | 78.96 | 86.12 | 86.13 | 86.1 |
|  | Image-only | 90.90 | 0.954 | 83.3 | 76.54 | 83.25 | 83.3 | 83.25 |
|  | RadEx-only | 77.67 | 0.887 | 76.37 | 63.67 | 76.4 | 76.4 | 76.35 |
| ResNeXt-50 | RadFuse | 93.24 | 0.964 | 87.18 | 80.5 | 87.07 | 87.2 | 87.17 |
|  | Image-only | 90.62 | 0.954 | 83.67 | 74.54 | 83.65 | 83.7 | 83.65 |
|  | RadEx-only | 78.78 | 0.889 | 77.7 | 65.94 | 77.68 | 77.7 | 77.7 |

To further evaluate the classification performance of our model on each DR severity level, we reported the F1, Recall, and Precision evaluation metrics per severity level in Table 3 and Fig.7. From the table, we observe that our model achieves exceptional accuracy in the Normal category, correctly identifying 99.0 of the cases. This high accuracy is crucial for reliable screening, ensuring that healthy patients are not misdiagnosed with DR. Similarly, the model demonstrates strong performance in the Moderate category, correctly classifying 88.02% of the instances, which is vital for initiating timely interventions. However, the recognition rates for the Severe and PDR categories are comparatively lower. Specifically, the model correctly identifies 71.67 of Mild cases, with a notable 23.33% misclassified as Moderate. For the Severe category, the correct classification rate drops to 63.64%, with 30.30% of cases misclassified as Moderate. The PDR category shows the lowest correct identification at 57.14, with 28.57% misclassified as Moderate and 14.29% as Severe. This pattern is consistent with findings in other studies, such as those by CLS[25], and MDGNet[13], where models also struggled with Severe and PDR cases as shown in Fig. 7a, and b.

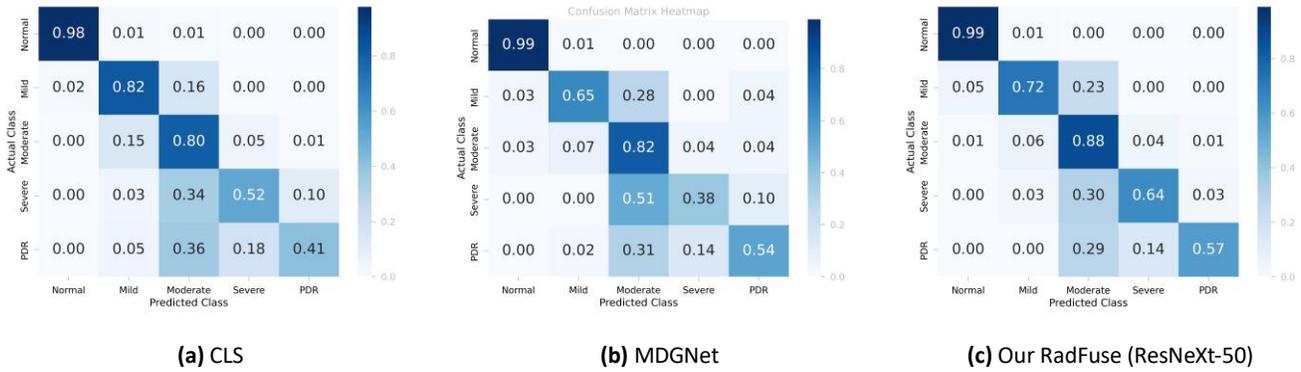

(a) CLS     (b) MDGNet     (c) Our RadFuse (ResNeXt-50)

**Figure 7.** Confusion matrices of three models on the APTOS-2019 dataset

**Table 3.** ResNeXt-50-based RadFuse Class-wise performance metrics on Aptos-2019 dataset

| Class | Recall (%) | Precision (%) | F1 Score (%) |
|---|---|---|---|
| Normal | 99.0 | 98.67 | 98.83 |
| Mild | 71.67 | 75.44 | 73.50 |
| Moderate | 88.02 | 79.46 | 83.52 |
| Severe | 63.64 | 60.00 | 61.76 |
| PDR | 57.14 | 90.32 | 70.0 |

To further verify the superiority of our results, we compare our best-performing ResNeXt-50-based RadFuse model with SOTA models for DR severity level grading. These include ADCNet[26], CLS[25], MIL-ViT[27], CANet[28], and Dual-Branch with Augmentation[29]. Moreover, we also included some SOTA generic models such as Swin[30], ViG*[31], and MDGNet[13]. All the comparison results are presented in Table 4. It should be noted, however, that differences in model architectures, methodologies, and experimental setups among these SOTA methods introduce variability in performance, which may limit the direct comparability to our results. Despite these variations, our proposed RadFuse consistently outperforms all these models by a significant margin. It can be seen that our proposed RadFuse outperforms all these models by a large margin. The crucial performance metric, QWK, reflects the superior reliability of our model, with a score of 93.24 %, which is higher than the 92.00 reported by MIL-ViT[27], 89.03 % from[29], and 90.0 % from CANet[28], showcasing RadFuse's consistent ability to grade DR severity. In the APTOS-2019 dataset, recall is an important metric due to the imbalanced distribution of severity levels between DRs. This demonstrates how accurately the model is able to identify true positive cases, especially in minority groups. Our ResNeXt-50-based RadFuse model achieved a recall of 87.2 %, outperforming all other models by at least 5 percentage points, indicating its effectiveness in detecting various DR levels, including underrepresented categories. In addition to the recall, RadFuse achieved the highest accuracy (87.07 %), F1-score (87.17 %), and AUC (0.964), underscoring its robustness in DR severity classification and its ability to balance precision and recall.

The lower performance in the Severe and PDR categories can be attributed to the limited number of training samples for these classes in the APTOS-2019 dataset. As highlighted in Table 1, the dataset contains only 136 training images for the Severe category—the smallest among all categories—which hampers the model's ability to learn distinctive features for accurate classification. Further analysis of the misclassifications in the figure below reveals that errors of the categories other than the PDR category are identified as neighboring categories. For instance, the MDGNet[13] discriminated 51% of the Sever category as Moderate and 28% of the Mild category as Moderate. The reason for this may be that the difference between the DR images of the neighboring categories is very small, which leads to the misidentification of all the models. This confusion is understandable, given that the clinical differences between adjacent DR stages can be minimal and difficult to discern, even for experienced clinicians. Compared to the models discussed in the literature, our method demonstrates superior performance in recognizing the Severe category. While MDGNet[13] achieved only 38% correct identification for Severe cases, our model correctly classifies 64%, indicating a significant improvement despite the small training samples for this category. This enhancement could be attributed to the inclusion of the non-linear Radex-based sinograms, which amplify critical features associated with advanced DR stages.

In addition to the multi-class severity grading, we evaluated the performance of our models on the binary classification task of distinguishing between DR and healthy cases. This task is critical for screening purposes, where the primary goal is to identify patients who require further ophthalmic examination. We trained and tested the same three CNN architectures—MobileNetV2, **Table 4.** Comparison with SOTA DR grading methods on APTOS-2019. The SOTA results were taken from[13,27,29]. The best results are in bold. The second best results are underlined.

| Method | Accuracy (%) | F1 (%) | QWK (%) | AUC | Precision (%) | Recall (%) |
|---|---|---|---|---|---|---|
| Swin[30] | 81.44 | 60.58 | - | 0.773 | 64.43 | 59.42 |
| ViG[31] | 82.81 | 66.48 | - | 0.80 | 69.42 | 64.52 |
| MDGNet[13] | 84.31 | 69.69 | - | 0.723 | 67.84 | - |
| CANet[28] | 83.2 | 81.3 | 90.0 | - | - | - |
| DANIL[32] | 83.8 | 67.2 | - | - | - | - |
| GREEN-ResNet50[33] | 84.4 | 83.6 | 90.8 | - | - | - |
| GREEN-SE-ResNext50[33] | 85.7 | 85.2 | 91.2 | - | - | - |
| ADCNet[26] | 83.40 | 67.02 | 74.78 | 0.9426 | 69.66 | 67.70 |
| CLS[25] | 84.36 | 70.49 | - | 0.938 | 70.49 | 70.51 |
| MIL-ViT[27] | 85.5 | 85.30 | 92.00 | 0.979 | - | - |
| Dual-Branch with Augmentation[29] | 83.17 | 82.64 | 89.03 | 0.898 | 82.66 | 83.17 |
| RadFuse (ResNeXt-50) | 87.07 | 87.18 | 93.24 | <u>0.964</u> | 87.17 | 87.2 |

VGG19, and ResNeXt-50— on a re-labeled version of the dataset for binary classification, differentiating between healthy (Grade 0) and DR (Grades 1–4). The key performance metrics of our RadFuse-based approach are presented in Table 5. As shown, RadFuse outperformed both image-only and RadEx-only models across all three CNNs. ResNeXt-50-based RadFuse achieved the best overall performance, with the highest QWK (98.17%), accuracy (99.09%) and F1-score (99.11%), indicating superior classification capability. MobileNetV2 and VGG19 also showed strong discriminative ability, with AUC values exceeding 0.996 and QWK scores over 0.97. The confusion matrices (Figure 8) offer insights into the classification outcomes. ResNeXt-50 demonstrated the fewest misclassifications, incorrectly predicting one DR case as healthy, and misclassifying four healthy cases as DR. VGG19 and MobileNetV2 had slightly higher error rates, with six and eight misclassifications, respectively. However, all models tended to misclassify healthy cases as DR, a trend that is acceptable in screening contexts, where minimizing missed diagnoses is crucial. This analysis further emphasizes that the models prioritize DR detection, which is essential in clinical screening to prevent undiagnosed DR progression. For a comprehensive evaluation, we also compared our best ResNeXt-50-based RadFuse model against recent SOTA methods for binary classification. As shown in Table 6, our approach outperformed all referenced methods, achieving the highest accuracy (99.09%) and recall (99.64%). Additionally, the high precision (98.58%) indicates a lower false-positive rate compared to other approaches, underscoring its effectiveness in screening applications.

**Table 5.** Detailed Performance Metrics of the Proposed Models for Aptos dataset Binary Classification

| Model | Configuration | Accuracy % | F1Score % | MCC % | QWK % | Precision % | Recall % |
|---|---|---|---|---|---|---|---|
| MobileNetV2 | RadFuse | 98.54 | 98.57 | 97.08 | 97.08 | 98.22 | 98.93 |

|  | Image-only | 97.79 | 97.63 | 95.60 | 95.59 | 97.7 | 97.57 |
|  | RadEx-only | 94.49 | 94.49 | 88.99 | 88.97 | 94.49 | 94.49 |
| VGG19 | RadFuse | 98.91 | 98.92 | 97.81 | 97.81 | 98.57 | 99.28 |
|  | Image-only | 97.24 | 97.24 | 94.51 | 94.48 | 97.24 | 97.24 |
|  | RadEx-only | 94.85 | 94.85 | 89.76 | 89.70 | 94.85 | 94.85 |
| ResNeXt-50 | RadFuse | 99.09 | 99.11 | 98.18 | 98.17 | 98.58 | 99.64 |
|  | Image-only | 97.43 | 97.43 | 94.85 | 94.85 | 97.43 | 97.43 |
|  | RadEx-only | 95.59 | 95.59 | 91.17 | 91.17 | 95.59 | 95.59 |

**Results on DDR Dataset**

**Table 6.** Comparison with SOTA Methods for Binary Classification. Results were taken from[29]

| Reference | Accuracy (%) | Precision (%) | Recall (%) |
|---|---|---|---|
| Blended features + DNN[34] | 96.10 | – | – |
| Hybrid model (VGG16 + Capsule network)[35] | 97.05 | – | – |
| CLS[25] | 98.36 |  | 98.37 |
| Dual-Branch with Augmentation[29] | 98.50 | 98.36 97.61 |  |
| ResNeXt-50 with RadFuse | 99.09 | 98.58 | 99.46 99.64 |

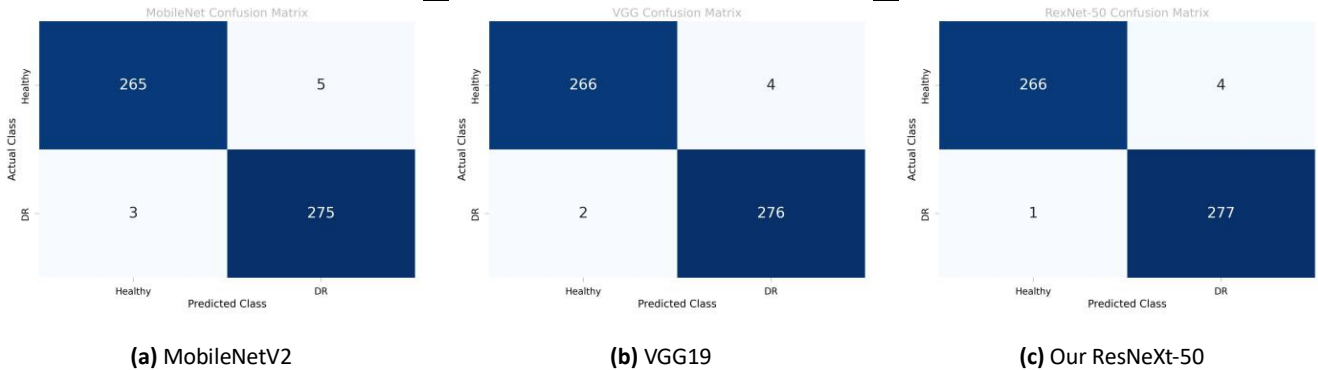

(a) MobileNetV2　　　　　　　　　(b) VGG19　　　　　　　　　(c) Our ResNeXt-50

**Figure 8.** Binary classification confusion matrices for our three different models on the APTOS-2019 dataset

To substantiate the generalizability of our RadFuse framework beyond the APTOS-2019 dataset, we conducted additional experiments using the DDR benchmark dataset. Table 7 presents the performance metrics of our RadFuse models compared to Image-only and RadEx-only configurations across three CNN architectures: ResNeXt-50, MobileNet, and VGG-19. With the ResNeXt-50 architecture, RadFuse achieved a QWK score of 85.50%, outperforming the image-only model's 81.00% by 4.50 percentage points. Key metrics including MCC, F1 Score, and AUC also improved, with the AUC rising from 0.921 to 0.948. For the VGG19 architecture, RadFuse demonstrated even substantial gains, achieving a QWK score of 84.82% and an AUC increase from 0.903 to 0.925. This trend continued with MobileNetV2, where RadFuse outperformed the image-only model in QWK (78.90% vs. 76.66%) and AUC (0.916 vs. 0.893), underscoring the effectiveness of integrating RadEx-transformed images with original fundus images within the RadFuse framework across various CNN architectures.

Notably, as shown in Figure 9, the reduction in misclassification rates for severe and PDR stages underscores the effciency of
RadFuse for enhancing their detection, as accurate detection of these stages is critical for timely intervention and prevention of vision loss. A detailed analysis of the confusion matrices for the ResNeXt-50 architecture on the DDR dataset highlights that the RadFuse framework significantly reduces misclassification rates in these stages compared to the image-only model.

Specifically, the image-only model exhibited a misclassification rate of 70.42% for the severe grade (50 out of 71 cases misclassified) and 22.55% for PDR (62 out of 275 cases misclassified). In contrast, the RadFuse model reduced these rates to 52.11% for the severe (37 out of 71 cases misclassified) and 16.36% for DR4 (45 out of 275 cases misclassified). This represents an 18.31% reduction in Severe grade misclassification and a 6.19% reduction in PDR grade misclassification, respectively. Similarly, the MobileNet architecture demonstrated a reduction in misclassification rates from 66.20% (47 out of 71 the severe cases misclassified) in the image-only model to 52.11% (37 out of 71 the severe cases misclassified) in the RadFuse model for the severe grade, and from 31.70% (87 out of 275 PDR cases misclassified) to 16.36% (45 out of 275 PDR cases misclassified) for PDR grade. For the VGG-19 architecture, the image-only model showed a severe grade misclassification rate of 57.75% (41 out of 71 cases misclassified), which was not reduced in the RadFuse model. Although this indicates a slight increase in DR3 misclassification for VGG-19, the overall performance metrics, including QWK and MCC, improved significantly, suggesting enhanced performance in other classes that compensate for this increase. These observations underscore the RadFuse framework's enhanced ability to accurately detect severe and PDR stages, which is paramount for early clinical interventions. The reduction in misclassification rates for these grades highlights RadFuse's efficacy in capturing complex, non-linear lesion patterns.

The per-class performance metrics in Table 8 illustrates the per-class evaluation metrics for the best performing image-only and RadFuse ResNeXt-50 models, highlighting improvements in the RadFuse configurations. The RadFuse models exhibit significant improvements in both AUC and Recall scores across all classes. RadFuse achieved an AUC of 0.973 for No DR, enhancing the image-only model's AUC of 0.96. More notably, for Mild, Moderate, Severe, and PDR, the RadFuse model achieved AUC scores of 0.904, 0.921, 0.958, and 0.987, respectively, compared to the image-only model's 0.806, 0.909, 0.857, and 0.968. In terms of Recall, the RadFuse model significantly improves sensitivity across all classes. For Mild and Moderate, recall rates increased from 77.12% to 82.36% and from 74.40% to 82.36%, respectively. Most importantly, for the severe DR stages, Severe recall improved from 29.58% to 47.89%, and PDR recall saw a marginal increase from 22.55% to 23.00%. While the improvement in PDR recall is modest, the substantial gain in Severe recall highlights RadFuse's enhanced capability to detect severe cases, which are critical for clinical interventions. Additionally, the F1 Score and Precision metrics also demonstrate consistent enhancements, reflecting a balanced improvement in both sensitivity and specificity.

To further validate the efficacy of our proposed RadFuse framework, we compared our best-performing ResNeXt-50based RadFuse model with several SOTA models for five-class DR severity level grading. Table 9 presents a comprehensive comparison. Our RadFuse model achieved an accuracy of 83.32% and a superior QWK of 85.50%, positioning it as the second-highest performer in accuracy and the top performer in QWK among the compared models. Compared to attention-based models like CABNet[36] and FA-Net[37], which achieved QWK scores of 78.63% and 82.68%, respectively, RadFuse not only matches but exceeds their performance, underscoring its reliability across all DR classes. These results suggest that while attention mechanisms contribute significantly to focusing on specific lesion types and enhancing grading performance data, the integration of RadEx-transformed features in RadFuse provides an additional layer of diagnostic information, further enhancing its discriminative capability. ViT+CSRA[38] achieves an accuracy of 82.35% but lacks the QWK metric, suggesting that while Vision Transformers are effective for DR grading, they may not capture fine lesion details as effectively as Radon-transformed representations or specialized attention models. In multimodal and hybrid models, strategies like DeepMT-DR[39] and FA+KCNet+R1[37] highlight the strengths of multitask and multimodal learning. DeepMT-DR achieved 83.60% accuracy and 80.20% QWK, illustrating the value of multitask learning in DR grading, while FA+KC-Net+R1 slightly outperformed with 83.96% accuracy and 84.76% QWK, showing the benefits of combining fine-grained attention with knowledge-based methods. Notably, RadFuse matches or slightly surpasses these complex models despite its relative simplicity, suggesting that non-linear Radon transformations as an added data representation enable effective capture of subtle lesion features similar to intricate multimodal systems.

**Table 7.** Performance Metrics on the DDR Dataset. Comparison of model performance for severity-level grading using different configurations. Best results are in bold. "Image-only" indicates models trained and tested using fundus images only, "RadEx-only" uses only RadEx-transformed images, and "RadFuse" indicates multi-representation models integrating both fundus images and RadEx-based sinograms.

| Model | Configuration | QWK % | MCC % | F1 Score % | AUC | Accuracy % | Sensitivity |
|---|---|---|---|---|---|---|---|
| MobileNetV2 | Image-only | 76.66 | 61.62 | 77.12 | 0.893 | 77.12 | 77.12 |
|  | RadEx-only | 43.55 | 24.64 | 57.73 | 0.743 | 57.73 | 57.73 |

|  | RadFuse | 78.90 | 65.90 | 79.30 | 0.915 | 79.3 | 79.3 |
|---|---|---|---|---|---|---|---|
|  | Image-only | 76.74 | 60.23 | 76.14 | 0.903 | 76.14 | 76.14 |
| VGG-19 | RadEx-only | 56.56 | 36.55 | 62.84 | 0.773 | 62.84 | 62.84 |
|  | RadFuse | 84.82 | 71.16 | 82.36 | 0.925 | 82.36 | 82.36 |
|  | Image-only | 81.00 | 69.67 | 81.51 | 0.921 | 81.50 | 81.50 |
| ResNeXt-50 | RadEx-only | 54.18 | 39.55 | 65.07 | 0.776 | 65.00 | 65.0 |
|  | RadFuse | 85.50 | 72.72 | 83.32 | 0.948 | 83.32 | 83.32 |

*Sensitivity Analysis of Augmentation Strategies*

We evaluated the effect of data augmentation strategies on the performance of RadFuse using the DDR dataset and the ResNeXt-50 architecture. Two approaches were compared: (1) joint augmentation of pre-generated RadEx-transformed images concatenated with fundus images and (2) dynamic recalculation of RadEx for each augmented version of the fundus images during training.

The results, detailed in Supplementary Table S1, demonstrate that RadFuse consistently outperformed both Image-only and RadEx-only models across both strategies, underscoring the value of integrating RadEx-transformed images with fundus images. Notably, recalculating RadEx for each augmentation improved performance, with QWK increasing from 85.0% to 86.0% and the F1 score rising from 83.0% to 85.0%. However, this improvement came at a significant computational cost, with training time per epoch increasing approximately 20-fold, from 3 minutes with joint augmentation to over an hour with recalculated augmentation. Considering this trade-off, we adopted the joint augmentation strategy for its computational efficiency. Future

|  | Normal | Mild | Moderate | Severe | PDR |
|---|---|---|---|---|---|
| Normal | 1851 | 14 | 15 | 0 | 0 |
| Mild | 64 | 75 | 48 | 0 | 2 |
| Moderate | 252 | 92 | 940 | 33 | 27 |
| Severe | 1 | 0 | 30 | 37 | 3 |
| PDR | 13 | 0 | 87 | 14 | 161 |

|  | Normal | Mild | Moderate | Severe | PDR |
|---|---|---|---|---|---|
| Normal | 1849 | 3 | 28 | 0 | 0 |
| Mild | 61 | 54 | 71 | 0 | 3 |
| Moderate | 236 | 73 | 965 | 28 | 42 |
| Severe | 0 | 0 | 31 | 34 | 6 |
| PDR | 5 | 1 | 33 | 6 | 230 |

(a) ResNeXt-50: Image-only model      (b) ResNeXt-50 : RadFuse model

**(c)** VGG19: Image-only model

**(d)** VGG19 : RadFuse model

**(e)** MobileNetV2: Image-only model

**(f)** MobileNetV2 : RadFuse model

**Figure 9.** Confusion matrices for severity grading for the Image-only and RadFuse models on the DDR dataset. work will focus on optimizing the RadEx calculation process to reduce its computational demands during dynamic image augmentation while further enhancing its performance benefits.

*Sensitivity Analysis of RadEx Transformation's Parameters*

We conducted a sensitivity analysis to determine optimal values for the RadEx transformation parameters $\Delta q$ and $c$. To evaluate the impact of step size $\Delta q$ on model performance, we assessed three step sizes: $\Delta q = \frac{M}{100}$, $\Delta q = \frac{M}{50}$, and $\Delta q = \frac{M}{25}$, where $M = 512$ pixels represents the image dimension (for a 512x512 pixel image). The number of curvature parameters $c$ was fixed at $\frac{M}{2}$. Table 10 summarizes the performance of the RadFuse and RadEx-only models at different $\Delta q$ values on the APTOS-2019 dataset. Results show that performance differences between step sizes $\Delta q = \frac{M}{50}$ and $\Delta q = \frac{M}{100}$ are minimal for both RadFuse and RadEx-only models. Specifically, for RadFuse, QWK scores range from 91.41% to 93.10%, MCC from 79.53% to 80.86%, and AUC from 0.955 to 0.964 across these step sizes. Similarly, RadEx-only exhibits QWK scores between 72.18% and 78.78%, MCC from 61.99% to 66.06%, and AUC from 88.38% to 88.88%. These minor variations suggest that both models are relatively robust within the tested $\Delta q$ range.

Among the evaluated step sizes, $\Delta q = \frac{M}{50}$ emerged as the optimal choice for both configurations. This setting achieves the best trade-off between performance metrics and computational cost, with RadFuse reaching peak values of QWK (93.10%), MCC (80.55%), and AUC (0.964), while RadEx-only attains its highest QWK and MCC at this step size. Additionally, this configuration shows the lowest misclassification rates for severe and PDR stages (Supplementary Figure S1). Misclassifications between severe and PDR (e.g., severe predicted as PDR and vice versa) remain consistent across Δq values, indicating that step size mainly affects misclassifications with less severe grades like DR2 (Supplementary Figure S1). Thus, Δq = $\frac{M}{50}$ provides balanced performance with minimal misclassification. Overall, the RadEx transformation demonstrates robustness to parameter selection within this range. While slight performance fluctuations are observed, they do not detract from overall efficacy, suggesting that while RadEx benefits from tuning Δq.

The sensitivity analysis on the curvature parameter $c$ also reveals that the RadFuse model performs reliably across different configurations, while careful tuning of $c$ can enhance specific performance metrics. We evaluated three $c$ settings as shown in Table 11 and Figure 10, where $M$ = 512 and Δq = $\frac{M}{2}$. At $c = \frac{M}{2}$, the RadFuse model achieves an optimal balance of performance and computational efficiency. Recall rates for Severe and PDR stages reach 51.52% and 54.00%, respectively, with the lowest rate of misclassifications among severe stages. This setting effectively captures essential features while avoiding major performance trade-offs, indicating that moderate adjustments to $c$ enhance classification accuracy. Increasing $c$ to $M$ yields further improvements, especially for PDR recall (68.00%), with slight increases in misclassifications for the Severe stage. This suggests that while performance gains are achievable with higher $c$ values, the model's stability remains strong across the tested range, and $c = \frac{M}{2}$ offers balanced solution with lower cost. Specifically, for the RadFuse model, $c = \frac{M}{2}$ achieved a balanced performance across all metrics, attaining the highest AUC of 0.964. Increasing $c$ to $M$ resulted in marginal improvements, with peak values for QWK (93.26%), MCC (82.81%), F1 Score (88.60%), and Accuracy (88.60%), though at increased computational cost. This configuration enhances the detection of specific features but does not universally improve severe stage performance, reinforcing $c = \frac{M}{2}$ as the optimal setting for balancing accuracy and efficiency. Detailed confusion matrices for these analyses are provided in Supplementary Figure S2.

**Limitations and future work**

Despite the promising results of RadFuse, some limitations remain. While the APTOS-2019 dataset is comprehensive, a broader evaluation across multiple datasets is necessary to confirm the generalizability of our approach. Although our model performed well on the imbalanced dataset, the class imbalance—particularly in the PDR stages—likely impacted the accuracy in these categories. Future work should focus on addressing this imbalance using advanced techniques, which could further improve the model's overall performance. Additionally, exploring alternative non-linear equations within the RadEx framework may capture more intricate retinal features, potentially enhancing feature extraction and classification accuracy. Moreover, while our optimized RadEx transform has proven effective in improving the feature representation of retinal images, its computational complexity could limit scalability, especially in large-scale or resource-constrained environments. By addressing these limitations in future research, we can enhance both the practical applicability and robustness of the RadFuse approach.

Table 8. Class-wise performance metrics on the DDR dataset for Image-only and RadFuse models (ResNeXt-50 architecture)

| Class | Image-only | | | | RadFuse | | | |
|---|---|---|---|---|---|---|---|---|
| | Recall % | Precision % | F1 Score % | AUC | Recall % | Precision % | F1 Score % | AUC |
| Normal | 98.29 | 84.54 | 90.90 | 0.960 | 98.35 | 85.96 | 91.74 | 0.973 |
| Mild | 21.16 | 32.52 | 25.64 | 0.806 | 28.57 | 41.22 | 33.75 | 0.904 |
| Moderate | 74.40 | 84.89 | 79.30 | 0.909 | 71.80 | 85.55 | 78.07 | 0.921 |
| Severe | 29.58 | 61.76 | 40.00 | 0.857 | 47.89 | 50.00 | 48.92 | 0.958 |
| PDR | 77.45 | 89.50 | 83.04 | 0.968 | 83.64 | 81.85 | 82.73 | 0.986 |

Table 9. Performance comparison on five-class DR grading on the DDR dataset of our RadFuse model against several SOTA models. Benchmark results are taken from[37] and[36]. Best results are in bold, and second-best results are underlined.

| Model | Accuracy % | QWK % | Remarks |
|---|---|---|---|
| ResNet50[36] | 75.57 | 74.27 | Standard CNN baseline without attention mechanisms |
| DenseNet-121[36] | 76.69 | 74.38 | Standard CNN baseline without attention mechanisms |
| CABNet[36] | 78.98 | 78.63 | Utilizes Category Attention Block (CAB) for imbalance handling |
| DeepMT-DR[39] | 83.60 | 80.20 | Employs a multitask approach to improve DR grading |
| ViT+CSRA[38] | 82.35 | - | Combines Vision Transformer with Class-Specific Representation Alignment (CSRA) for feature learning |
| FA-Net[37] | 82.10 | 82.68 | Fine-grained attention modules, strong performance on five-class grading |
| FA+KC-Net+R1[37] | 83.96 | <u>84.76</u> | Fusion of fine-grained attention and knowledge-based network |
| RadFuse (ResNeXt-50) | <u>83.32</u> | 85.50 | RadFuse with non-linear Radon transformations for enhanced lesion focus |

**Table 10.** Performance comparison of RadFuse and RadEx-only models across different step sizes $\Delta q$

| Step Size $\Delta q$ | Model | QWK % | MCC % | F1 % | Accuracy % | AUC | Recall % | Precision % |
|---|---|---|---|---|---|---|---|---|
| $\frac{M}{25}$ | RadFuse | <u>91.51</u> | 79.53 | 86.36 | 86.36 | <u>0.957</u> | 86.36 | 86.36 |
| | RadEx-only | 72.18 | 61.99 | 75.44 | 75.44 | 0.883 | 75.44 | 75.44 |
| $\frac{M}{50}$ | RadFuse | 93.10 | 80.55 | 87.16 | 87.16 | 0.964 | 87.16 | 87.16 |
| | RadEx-only | 78.78 | 65.93 | 77.69 | 77.69 | 0.888 | 77.69 | 77.69 |
| $\frac{M}{100}$ | RadFuse | 91.41 | 80.86 | 87.32 | 87.32 | 0.955 | 87.32 | 87.32 |
| | RadEx-only | 75.97 | 66.06 | 77.69 | 77.69 | 0.884 | 77.69 | 77.69 |

## Related work

Deep learning techniques have become increasingly popular for DR detection and grading, outperforming traditional machine learning methods in accuracy and effectiveness[29]. Early work by Gulshan et al.[5] and Ting et al.[6] demonstrated that CNN could surpass general ophthalmologists in detecting referable DR. Subsequent studies aimed to refine DR grading granularity. Several researchers have developed various methods for the diagnosis of DR using different approaches. For example, Pratt et al.[7] used CNNs with data augmentation to improve model generalization, while Shanthi et al.[40] enhanced the AlexNet architecture for better grading accuracy. More recent developments include Islam et al.'s supervised contrastive learning (SCL) approach, which significantly improved binary classification and multi-stage DR grading using the APTOS 2019 dataset[25]. The SCL model attained an accuracy of 98.36% and an AUC of 0.985 for binary classification, and 84.36% accuracy for five-stage grading. Hossein et al.[29] also leveraged transfer learning with ResNet50 and EfficientNetB0 to address class imbalances, utilizing data from multiple datasets. More recently, MDGNet introduced a novel architecture that integrates local and global lesion features for improved DR grading[13]. Despite these advancements, accurately capturing early-stage DR features, such as microaneurysms and small hemorrhages, remains challenging due to their subtle nature and variability[41]. CNN-based models often struggle with capturing subtle and non-linear in DR-related lesions, making it difficult to differentiate between adjacent severity levels.

Image transformation techniques are essential in medical imaging, facilitating effective feature extraction crucial for precise diagnostics[42]. The Radon transform is a mathematical integral transform that computes projections of an image along specified directions. It has been widely used in computed tomography for image reconstruction from projection data[15]. The transform effectively captures linear features and has been applied in medical image analysis for tasks such lung and breast cancer **Table 11.** Performance comparison of ResNeXt-50-based RadFuse and RadEx-only models across varying curvature parameter settings *c*. The best model in each configuration (RadFuse and RadEx-only) is highlighted in bold, and the second best is underlined.

| $c$ | Model | QWK % | MCC % | F1 % | Accuracy % | AUC | Recall % | Precision % |
|---|---|---|---|---|---|---|---|---|
| $M_4$ | RadFuse | 91.98 | 80.26 | 87.00 | 86.99 | 0.952 | 86.99 | 86.99 |
| | RadEx-only | 71.03 | 63.88 | 76.57 | 76.56 | 0.889 | 76.56 | 76.56 |
| $M_2$ | RadFuse | 93.10 | 80.55 | 87.16 | 87.16 | 0.964 | 87.16 | 87.16 |
| | RadEx-only | 78.78 | 65.93 | 77.69 | 77.68 | 0.888 | 77.69 | 77.69 |
| $N$ | RadFuse | 93.26 | 82.81 | 88.60 | 88.60 | 0.960 | 88.60 | 88.60 |
| | RadEx-only | 73.49 | 63.88 | 76.24 | 76.24 | 0.893 | 76.24 | 76.24 |

classification[16,43]. Its key benefit is its ability to simplify a complex image structure into analyzable projections, essential for both image reconstruction and feature extraction. Its integral nature enables it to accumulate all image data along specified lines through an object, highlighting critical structures and features for disease detection. TAVAKOLI et al.[12] demonstrated the Radon transform effectiveness in enhancing microaneurysm detection in retinal images. It is worth mentioning that these studies primarily focused on using extracted features from the sinogram rather than directly utilizing the sinogram images for classification tasks. However, the linear Radon transform struggles to capture non-linear features such as curved edges or irregular textures, typical in pathological conditions. Its linear assumptions may not accurately represent these complex features. Therefore, Non-linear extensions of the Radon transform have been proposed to capture more complex patterns in image data. Our previous work introduced the RadEx transform, a customized non-linear Radon transformation designed to improve feature extraction in chest X-ray images for COVID-19 detection[17]. The RadEx transform demonstrated the ability to highlight non-linear features not readily apparent in the spatial domain, suggesting its potential applicability to other medical imaging tasks. However, the application of non-linear Radon transformations for DR grading remains relatively unexplored. Therefore, building on the concept of the Radex transform, this study explores the use of an optimized version of RadEx, for enhanced feature extraction from fundus images in the context of DR grading.

Combining multiple imaging modalities can provide complementary information, leading to improved diagnostic performance. multimodal deep learning models integrate data from different sources to enhance feature representation and capture complex patterns[24]. In the context of DR, multimodal approaches have been less common due to the reliance on retinal fundus images as the primary data source. Some studies have explored the use of optical coherence tomography (OCT) alongside fundus images to improve DR detection[19]. However, OCT imaging is not always readily available, especially in resource-limited settings. Our work builds upon these insights by introducing a new multi-representation deep learning framework that integrates non-linear RadEx transformations-based sinogram images with retinal fundus images. This approach enhances feature extraction and improves DR severity grading accuracy.

## Conclusion

In this study, we introduced RadFuse, a multi-representation deep learning approach designed to enhance DR detection and severity grading by integrating an optimized non-linear RadEx transformation with original fundus images. Our method significantly improves CNN performance in detecting and classifying DR severity levels. Extensive experiments on the APTOS2019 and DDR datasets demonstrate that RadFuse multi-representation consistently outperforms traditional image-only models and several SOTA methods. By capturing complex non-linear patterns and distributed retinal lesions, the RadEx-sinogram provides valuable discriminative information not easily visible in the original images. These findings highlight the potential of combining spatial and transformed domain information to improve DR detection and grading.

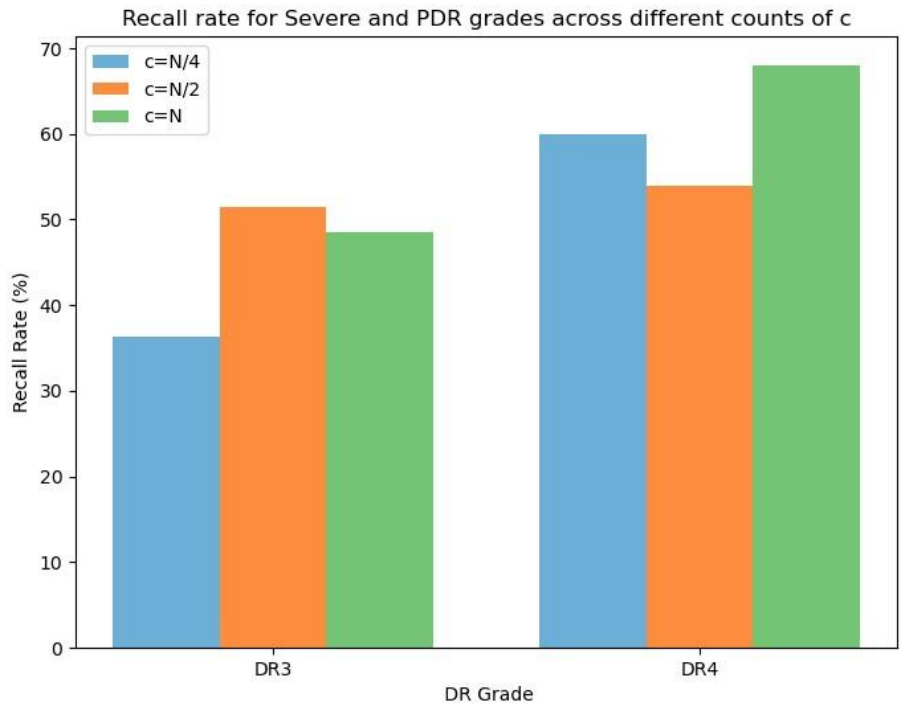

**Figure 10.** Recall rates for severe and PDR grades across different curvature parameter configurations in the ResNeXt-50-based RadFuse model. The bar chart shows the specific recall rates for each grade under three configurations of the curvature parameter *c*.

## References

1. Federation, I. D. *IDF Diabetes Atlas, 9th ed.* (International Diabetes Federation, 2019).
2. Thanikachalam, V., Kabilan, K. & Erramchetty, S. Optimized deep cnn for detection and classification of diabetic retinopathy and diabetic macular edema. *BMC Med. Imaging* 24, 227, DOI: 10.1186/s12880-024-01406-1 (2024).
3. Kusakunniran, W. *et al.* Detecting and staging diabetic retinopathy in retinal images using multi-branch cnn. *Appl. Comput. Informatics* (2022).
4. Abràmoff, M. D. *et al.* Automated analysis of retinal images for detection of referable diabetic retinopathy. *JAMA ophthalmology* 131, 351–357 (2013).
5. Gulshan, V. *et al.* Development and validation of a deep learning algorithm for detection of diabetic retinopathy in retinal fundus photographs. *jama* 316, 2402–2410 (2016).
6. Ting, D. S. W. *et al.* Development and validation of a deep learning system for diabetic retinopathy and related eye diseases using retinal images from multiethnic populations with diabetes. *Jama* 318, 2211–2223 (2017).
7. Pratt, H., Coenen, F., Broadbent, D. M., Harding, S. P. & Zheng, Y. Convolutional neural networks for diabetic retinopathy. *Procedia computer science* 90, 200–205 (2016).
8. Gadekallu, T. R. *et al.* Early detection of diabetic retinopathy using pca-firefly based deep learning model. *Electronics* 9, 274 (2020).
9. Seoud, L., Hurtut, T., Chelbi, J., Cheriet, F. & Langlois, J. P. Red lesion detection using dynamic shape features for diabetic retinopathy screening. *IEEE transactions on medical imaging* 35, 1116–1126 (2015).

10. De Fauw, J. *et al.* Clinically applicable deep learning for diagnosis and referral in retinal disease. *Nat. medicine* 24, 1342–1350 (2018).

11. Bala, R., Sharma, A. & Goel, N. Ctnet: convolutional transformer network for diabetic retinopathy classification. *Neural Comput. Appl.* 36, 4787–4809 (2024).

12. Tavakoli, M. *et al.* Automated microaneurysms detection in retinal images using radon transform and supervised learning: application to mass screening of diabetic retinopathy. *IEEE Access* 9, 67302–67314 (2021).

13. Wang, Y., Wang, L., Guo, Z., Song, S. & Li, Y. A graph convolutional network with dynamic weight fusion of multi-scale local features for diabetic retinopathy grading. *Sci. Reports* 14, 5791 (2024).

14. Toft, P. The radon transform. *Theory Implementation (Ph. D. Diss. Tech. Univ. Denmark)* (1996).

15. Deans, S. R. *The Radon transform and some of its applications* (Courier Corporation, 2007).

16. Raaj, R. S. Breast cancer detection and diagnosis using hybrid deep learning architecture. *Biomed. Signal Process. Control.* 82, 104558 (2023).

17. Islam, A., Mohsen, F., Shah, Z. & Belhaouari, S. B. Introducing novel radon based transform for disease detection from chest x-ray images. In *2024 6th International Conference on Pattern Analysis and Intelligent Systems (PAIS)*, 1–5 (IEEE, 2024).

18. (APTOS), A. P. T.-O. S. Messidor-adcis. https://www.kaggle.com/c/aptos2019-blindness-detection (2019). Accessed: Sep. 11, 2021. [Online].

19. Li, T. *et al.* Diagnostic assessment of deep learning algorithms for diabetic retinopathy screening. *Inf. Sci.* 501, 511–522 (2019).

20. Graham, B. Kaggle diabetic retinopathy detection competition report (2015). Available at: https://www.kaggle.com/c/diabetic-retinopathy-detection/discussion/15801.

21. Xie, S., Girshick, R., Dollár, P., Tu, Z. & He, K. Aggregated residual transformations for deep neural networks. In *Proceedings of the IEEE conference on computer vision and pattern recognition*, 1492–1500 (2017).

22. Sandler, M., Howard, A., Zhu, M., Zhmoginov, A. & Chen, L.-C. Mobilenetv2: Inverted residuals and linear bottlenecks. In *Proceedings of the IEEE conference on computer vision and pattern recognition*, 4510–4520 (2018).

23. Simonyan, K. & Zisserman, A. Very deep convolutional networks for large-scale image recognition. *arXiv preprint arXiv:1409.1556* (2014).

24. Mohsen, F., Ali, H., El Hajj, N. & Shah, Z. Artificial intelligence-based methods for fusion of electronic health records and imaging data. *Sci. Reports* 12, 17981 (2022).

25. Islam, M. R. *et al.* Applying supervised contrastive learning for the detection of diabetic retinopathy and its severity levels from fundus images. *Comput. biology medicine* 146, 105602 (2022).

26. Yue, G. *et al.* Attention-driven cascaded network for diabetic retinopathy grading from fundus images. *Biomed. Signal Process. Control.* 80, 104370 (2023).

27. Bi, Q. *et al.* Mil-vit: A multiple instance vision transformer for fundus image classification. *J. Vis. Commun. Image Represent.* 97, 103956, DOI: https://doi.org/10.1016/j.jvcir.2023.103956 (2023).

28. Li, X. *et al.* Canet: cross-disease attention network for joint diabetic retinopathy and diabetic macular edema grading. *IEEE transactions on medical imaging* 39, 1483–1493 (2019).

29. Shakibania, H., Raoufi, S., Pourafkham, B., Khotanlou, H. & Mansoorizadeh, M. Dual branch deep learning network for detection and stage grading of diabetic retinopathy. *Biomed. Signal Process. Control.* 93, 106168 (2024).

30. Liu, Z. *et al.* Swin transformer: Hierarchical vision transformer using shifted windows. In *Proceedings of the IEEE/CVF international conference on computer vision*, 10012–10022 (2021).

31. Han, K., Wang, Y., Guo, J., Tang, Y. & Wu, E. Vision gnn: An image is worth graph of nodes. *Adv. neural information processing systems* 35, 8291–8303 (2022).

32. Gong, L., Ma, K. & Zheng, Y. Distractor-aware neuron intrinsic learning for generic 2d medical image classifications. In *Medical Image Computing and Computer Assisted Intervention–MICCAI 2020: 23rd International Conference, Lima, Peru, October 4–8, 2020, Proceedings, Part II 23*, 591–601 (Springer, 2020).

33. Liu, S., Gong, L., Ma, K. & Zheng, Y. Green: a graph residual re-ranking network for grading diabetic retinopathy. In *Medical Image Computing and Computer Assisted Intervention–MICCAI 2020: 23rd International Conference, Lima, Peru, October 4–8, 2020, Proceedings, Part V 23*, 585–594 (Springer, 2020).

34. Bodapati, J. D. *et al.* Blended multi-modal deep convnet features for diabetic retinopathy severity prediction. *Electronics* 9, 914 (2020).

35. Kumar, G., Chatterjee, S. & Chattopadhyay, C. Dristi: a hybrid deep neural network for diabetic retinopathy diagnosis. *Signal, Image Video Process.* 15, 1679–1686 (2021).

36. He, A., Li, T., Li, N., Wang, K. & Fu, H. Cabnet: Category attention block for imbalanced diabetic retinopathy grading. *IEEE Transactions on Med. Imaging* 40, 143–153 (2020).

37. Tian, M. *et al.* Fine-grained attention & knowledge-based collaborative network for diabetic retinopathy grading. *Heliyon* 9 (2023).

38. Gu, Z. *et al.* Classification of diabetic retinopathy severity in fundus images using the vision transformer and residual attention. *Comput. Intell. Neurosci.* 2023, 1305583 (2023).

39. Wang, X. *et al.* Joint learning of multi-level tasks for diabetic retinopathy grading on low-resolution fundus images. *IEEE J. Biomed. Heal. Informatics* 26, 2216–2227 (2021).

40. Shanthi, T. & Sabeenian, R. Modified alexnet architecture for classification of diabetic retinopathy images. *Comput. & Electr. Eng.* 76, 56–64 (2019).

41. Lam, C., Yu, C., Huang, L. & Rubin, D. Retinal lesion detection with deep learning using image patches. *Investig. ophthalmology & visual science* 59, 590–596 (2018).

42. Arulmurugan, R. & Anandakumar, H. Early detection of lung cancer using wavelet feature descriptor and feed forward back propagation neural networks classifier. In *Computational vision and bio inspired computing*, 103–110 (Springer, 2018).

43. Rani, K. V., Sumathy, G., Shoba, L., Shermila, P. J. & Prince, M. E. Radon transform-based improved single seeded region growing segmentation for lung cancer detection using ampwsvm classification approach. *Signal, Image Video Process.* 17, 4571–4580 (2023).

## Acknowledgements
Open Access funding provided by the Qatar National Library.

## Competing interests
The authors declare no competing interests.

## Author contributions statement
F.M. and S.B. contributed to the conceptualization of the study. F.M. performed the simulations and contributed to the writing of the original manuscript. S.B. and Z.S. analyzed the results. S.B. and Z.S. revised the manuscript. S.B. and Z.S. supervised the study. All authors read and approved the final manuscript.

## Data Availability
The APTOS dataset is openly available at: https://www.kaggle.com/competitions/aptos2019-blindness-detection (accessed on
12 August 2024). DDR dataset is available from https://github.com/nkicsl/DDR-dataset( accessed on 15 November 2024.)